\documentclass[letterpaper, 10 pt, conference]{ieeeconf}  % Comment this line out if you need a4paper

\IEEEoverridecommandlockouts                              % This command is only needed if 
\usepackage{times} % assumes new font selection scheme installed
\usepackage{amsmath} % assumes amsmath package installed
\usepackage{amssymb}  % assumes amsmath package installed
\usepackage{xcolor} 

\usepackage{booktabs, multirow} 
\usepackage{float}
\usepackage{graphicx}
\usepackage{hyperref}
\usepackage{url}
\usepackage{subcaption}

\title{\LARGE \bf
FalconGym: A Photorealistic Simulation Framework for Zero-Shot Sim-to-Real Vision-Based Quadrotor Navigation
}

\author{Yan Miao$^1$, Will Shen$^1$ and Sayan Mitra$^1$
\thanks{$^1$Department of Electrical and Computer Engineering, University of Illinois at Urbana-Champaign, Urbana, IL, 61801, USA} 
}

\begin{document}

\maketitle
\thispagestyle{empty}
\pagestyle{empty}

%%%%%%%%%%%%%%%%%%%%%%%%%%%%%%%%%%%%%%%%%%%%%%%%%%%%%%%%%%%%%%%%%%%%%%%%%%%%%%%%
\begin{abstract}

% We address the task of enabling a vision-based quadrotor to autonomously traverse multiple gate configurations under a zero-shot sim-to-real setup. 
We present a novel framework demonstrating zero-shot sim-to-real transfer of visual control policies learned in a Neural Radiance Field (NeRF) environment for quadrotors to fly through racing gates.
Robust transfer from simulation to real flight poses a major challenge, as standard simulators often lack sufficient visual fidelity. 
To address this, we construct a photorealistic simulation environment of quadrotor racing tracks, called \emph{FalconGym}, which provides effectively unlimited synthetic images for training.
Within FalconGym, we develop a pipelined approach for crossing gates that combines (i) a Neural Pose Estimator (NPE) coupled with a Kalman filter to reliably infer quadrotor poses from single-frame RGB images and IMU data, and (ii) a self-attention-based multi-modal controller that adaptively integrates visual features and pose estimation. 
This multi-modal design compensates for perception noise and intermittent gate visibility.
We train this controller purely in FalconGym with imitation learning and deploy the resulting policy to real hardware with no additional fine-tuning.
Simulation experiments on three distinct tracks (circle, U-turn and figure-8) demonstrate that our controller outperforms a vision-only state-of-the-art baseline in both success rate and gate-crossing accuracy.
In 30 live hardware flights spanning three tracks and 120 gates, our controller achieves a 95.8\% success rate and an average error of just 10\,cm when flying through 38\,cm-radius gates.
% thereby conforming to hard constraints with minimal violations.

% Keywords: nerf2real, sensor fusion, drone, 
\end{abstract}

%%%%%%%%%%%%%%%%%%%%%%%%%%%%%%%%%%%%%%%%%%%%%%%%%%%%%%%%%%%%%%%%%%%%%%%%%%%%%%%%
\section{INTRODUCTION}

Autonomous vision-based quadrotor control is crucial for a range of high-impact applications, including racing, search-and-rescue, and exploration. 
Several challenges exist for visual flight. 
First, visual control is often fragile: Geles~et~al.~\cite{DBLP:conf/rss/GelesBRX024} demonstrated impressive quadrotor racing performance with a vision-only actor-critic approach, yet their system degraded when gates were out of view for multiple frames, posing safety risks.
Second, sim-to-real transfer remains difficult \cite{Zhao2020SimtoRealTI} because visual control policies trained in conventional simulators may not generalize reliably to the real world due to differences in visual features and dynamics.

\begin{figure}[htbp]
    \centering
    \includegraphics[width=\linewidth]{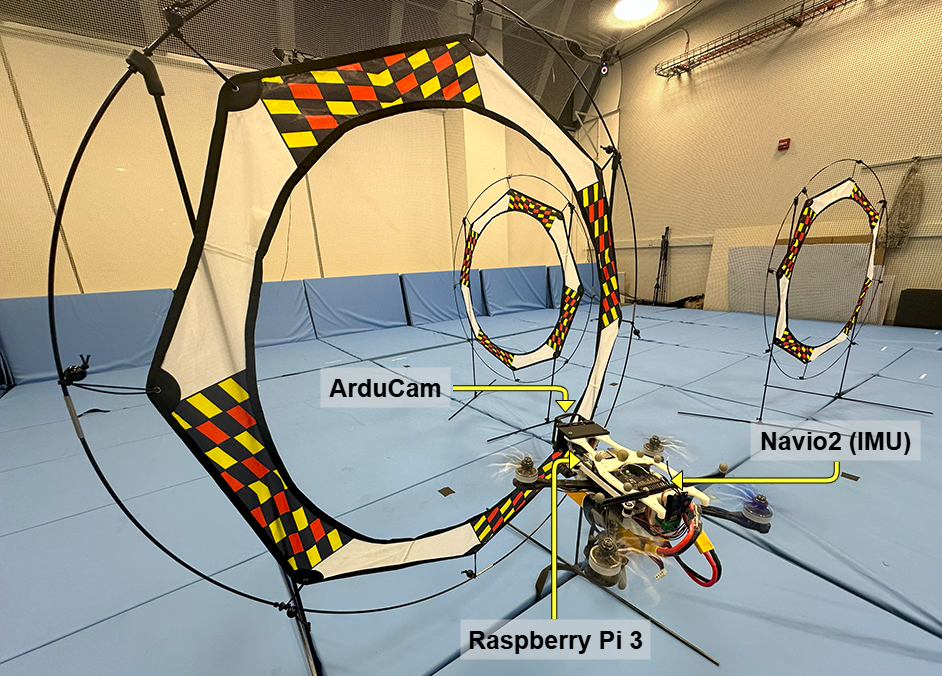}
    \caption{\small{Our autonomous quadrotor, equipped with an ArduCam RGB camera and an IMU-enabled Navio2 on a Raspberry Pi 3, can successfully fly through a sequence of small 38 cm-radius gates with an average of only 10 cm error to gates center.}}
    \label{fig:drone_gates}
\end{figure}

Recent advances in neural scene representations, such as Neural Radiance Fields (NeRF)~\cite{10.1145/3503250} and Gaussian Splatting~\cite{kerbl3Dgaussians}, 
% , and signed-distance functions~\cite{Osher2003}, 
provide an opportunity for creating photorealistic simulation environments, which can be used for training visual control policies. 
Although prior works have shown promising zero-shot neural-scene to real transfer for humanoid robots~\cite{Byravan2022NeRF2RealST} and grasping~\cite{torne2024rialto}, few have focused on quadrotors, which demand higher speeds and control frequencies than ground robots. 
SOUS VIDE~\cite{low2024sousvide} is a notable exception, which uses Gaussian Splat \cite{kerbl3Dgaussians} for neural scene representation and achieves impressive zero-shot drone navigation by distilling an expert MPC controller to a lightweight end-to-end policy.
% but its navigation trajectories involve relatively simple paths and do not include sharp turns.

In this paper, we address the sim-to-real gap for quadrotor gate crossing by developing a photorealistic simulation environment, \emph{FalconGym}, and designing a new visual control model for quadrotors using it.
Within FalconGym, we design a Neural Pose Estimator (NPE), which combined with a Kalman filter, to produce pose estimates. 
We then propose a self-attention-based multi-modal control policy that, unlike conventional fusion, can dynamically reweight the vision features and pose estimation inputs based on gate visibility to enhance robustness against perception errors. 
Finally, we train our multi-modal controller in FalconGym via imitation learning and deploy the resulting policy on a real quadrotor without any additional fine-tuning. 
In FalconGym, our controller completes ten laps on each of the three four-gate tracks (circle, U-turn, and figure-8) with a 100\% success rate across all tracks and an average gate-crossing error of 6.3, 10.1, and 5.1 cm for 38 cm-radius gates. 
In real-world experiments using the same tracks and lap count, the success rate and precision decrease slightly but remain notable: 100\%, 87.5\%, and 100\% success on the circle, U-turn, and figure-8 tracks with average errors of 10.7, 11.1, and 9.4cm. This performance is impressive given the relatively large size of our quadrotor (30 cm × 34 cm × 11 cm), which leaves little margin for error for 38 cm-radius gates.

In summary, our contribution is threefold. First, we demonstrate a novel framework to achieve zero-shot sim-to-real transfer of NeRF-trained visual control policies for quadrotors navigating through a sequence of small gates. Second, we develop \emph{FalconGym}, a photorealistic simulation environment for agile, vision-based quadrotor flight, providing an open benchmark for visual flight control research. Third, we make two innovations that make our neural controller's architecture practical:(i) A one-shot \emph{Neural Pose Estimator (NPE)} coupled with a Kalman filter; (ii) A self-attention-based sensor fusion scheme that integrates vision features and pose estimation for quadrotor flight.
% (ii) A quantile-regression approach for dataset augmentation, that enhances resilience to perception errors;

\section{RELATED WORK}

\paragraph{Neural Scene Representations in Robotics}
Neural scene representations like NeRF~\cite{10.1145/3503250} and Gaussian Splat \cite{kerbl3Dgaussians} have recently emerged as powerful tools for photorealistic 3D scene reconstruction. 
Variants and extensions have been applied to diverse robotics tasks, including 3D scene editing~\cite{ni2025efficientinteractive3dmultiobject}, pose estimation~\cite{yen2020inerf}, grasping~\cite{Dai2023GraspNeRF} and navigation~\cite{nerf-nav}.
% , and grasping~\cite{Dai2023GraspNeRF}. 
% 
% While these approaches exploit NeRF’s ability to synthesize novel, high-fidelity viewpoints, few works have addressed its capability to transfer to real world. 
% % 
% Notable exceptions include 
NeRF2Real~\cite{Byravan2022NeRF2RealST} and RialTo~\cite{torne2024rialto} demonstrate the potential sim-to-real transfer for ground robots and robot arm manipulation. 
\cite{low2024sousvide} achieves zero-shot drone navigation using policies trained in Gaussian splat scenes.
In addition to using NeRF representations for training, our work also uses a completely different solution architecture than \cite{low2024sousvide} and we solve the navigation problem with harder constraints in the shape of 38 cm-radius gates. 
% However, their work focuses more on quadrotors following user-defined trajectories whereas our approach focuses on the hard-constrained task of crossing narrow 38 cm-radius gates.

\paragraph{Sim-to-Real Transfer in Robotics}
Sim-to-real transfer has long been a cornerstone of robotics research, valued for its efficient and risk-free training. 
Traditional methods such as domain randomization~\cite{10.1109/IROS.2017.8202133} and system identification~\cite{10.5555/59617} are commonly employed to narrow the gap between simulation and reality. 
These techniques and their variations have enabled successful sim-to-real deployments in quadruped locomotion~\cite{doi:10.1126/scirobotics.abc5986}, bipedal walking~\cite{10.1109/IROS40897.2019.8968053}, and autonomous driving~\cite{hu2023simulationhelpsautonomousdrivinga}. 
% robotic grasping~\cite{10.1109/ICRA.2018.8460875}, 
% With the rise of NeRF-based simulations, there is growing potential to further enhance visual realism, thus improving transfer success for vision-centric tasks like ours.

\paragraph{Visual Flight}
Visual flight has attracted considerable interest recently. 
Kaufmann~et~al.~\cite{kaufmann2018deep} combined convolutional neural networks with advanced path-planning to achieve drone racing capabilities, while Geles~et~al. \cite{DBLP:conf/rss/GelesBRX024} extended agile flight racing to an end-to-end image-based policy, with an impressive speed upto 40 km/h. However, their visual control's performance deteriorates when racing gates are out of view for extended periods.
Other approaches have explored learning aggressive acrobatic flying maneuvers through imitation learning~\cite{kaufmann2020RSS} or through carefully simulated sensor noise for robust outdoor flight~\cite{doi:10.1126/scirobotics.abg5810}. 
Further progress has been made in autonomous flight to match human champion performance with Visual Inertial Odometry (VIO) and deep reinforcement learning with onboard sensors~\cite{kaufmann2023champion}. 
% Nevertheless, as drones increase in complexity and speed, accurate and robust perception remains critical—particularly for tasks where gates may be partially or intermittently visible.

\paragraph{Sensor Fusion}
Robots frequently integrate multiple sensing modalities to improve accuracy and resilience. 
In prior work, ~\cite{low2024sousvide}~\cite{kaufmann2020RSS} combine IMU measurements with visual features through a MLP for flight control. 
Recent sensor fusion approaches have also embraced attention-based architectures~\cite{10.5555/3295222.3295349}, including fusing camera and LiDAR signals for 3D perception~\cite{10188840}, or employing self-attention on electromyography (EMG) and IMU data for lower-limb locomotion control~\cite{10584554}, fusing goal image with current observation through transformers for better navigation \cite{shah2023vint}, However, most of these advances have been focused on ground robots rather than aerial systems.
% fusing Radar with camera using cross-attention mechanism for Waymo self-driving \cite{10.1007/978-3-031-19839-7_23}.

% We draw inspiration from recent advances in NeRF-based rendering, sim-to-real transfer, vision-based flight, and sensor fusion. By uniting these threads into a single framework, we enable zero-shot sim-to-real quadrotor navigation through gates, offering both high-fidelity simulation for training and reliable sensor fusion for robust in-flight performance.

\begin{figure}[htbp]
    \centering
    \includegraphics[width=\linewidth]{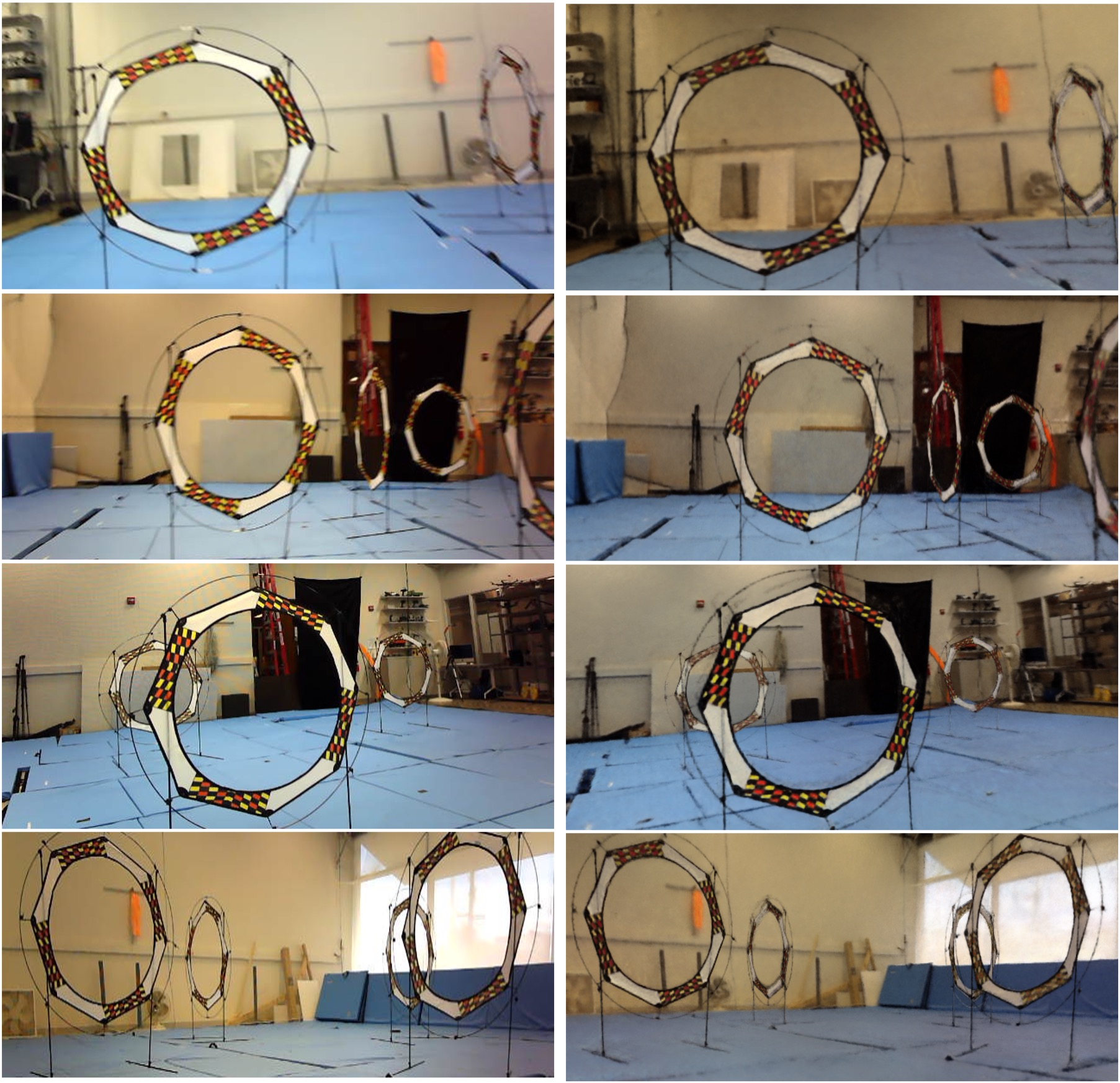}
    \caption{\small{FalconGym can render photorealistic images from different poses. The left column shows four real world images captured by the Arducam in our flight area, while the right column displays images rendered by FalconGym at the same coordinates.}
}
    \label{fig:real_vs_sim}
\end{figure}

\section{METHODOLOGY}

In this paper, we address the challenge of enabling a vision-centric quadrotor to autonomously fly through a series of small 38 cm-radius gates while adhering to hard safety constraints—specifically, avoiding collisions and staying on track, as shown in Figure \ref{fig:drone_gates} and Figure \ref{fig:trajactory_plot}.
Our objective is to train the controller entirely in simulation and zero-shot transfer the resulting control policy directly to real-world environments without additional fine-tuning. 
% leveraging NeRF \cite{10.1145/3503250} for high-fidelity visual realism,

\begin{figure*}[htbp]
    \centering
    \includegraphics[width=\linewidth]{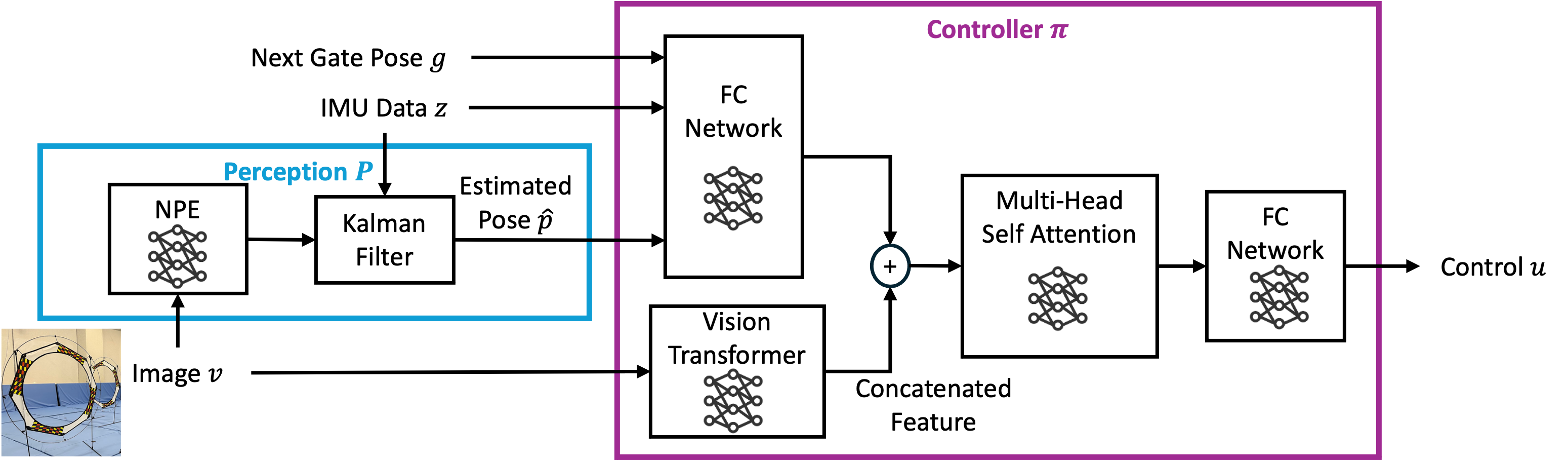}
    \caption{\small{The architecture of our visual flight system consists of a perception module and a control module. In the blue box, the perception module processes onboard camera images using a ViT-based Neural Pose Estimator (NPE) and improves its estimates with IMU data via a Kalman filter. The purple box details the control module, where a self-attention mechanism fuses state embeddings with vision features.}}
    \label{fig:perception-control-architecture}
\end{figure*}

In the following subsections, we first describe how we construct our \emph{FalconGym} simulation environments that are aligned with the real-world coordinate system.
We then introduce our perception module, which leverages a Neural Pose Estimator (NPE) to infer the quadrotor’s state from onboard camera images, where its estimate is further improved by fusing high-frequency IMU data via a Kalman filter~\cite{kalman1960new}. 
Lastly, we detail our controller design, which fuses visual features and pose estimation with a self-attention mechanism and is trained via DAGGER-style \cite{pmlr-v15-ross11a} imitation learning.

\subsection{FalconGym Construction}
\label{sec:nerf-construction}
% High-fidelity simulation is crucial in vision-based quadrotor navigation, as realistic visual cues can significantly enhance sim-to-real transfer. 
% 
% Recent advances in NeRF \cite{10.1145/3503250} enable photorealistic rendering of complex scenes from arbitrary viewpoints, making it attractive simulation environment for vision-based flying through gates tasks.

\begin{figure*}[htbp]
    \centering
    \includegraphics[width=\linewidth]{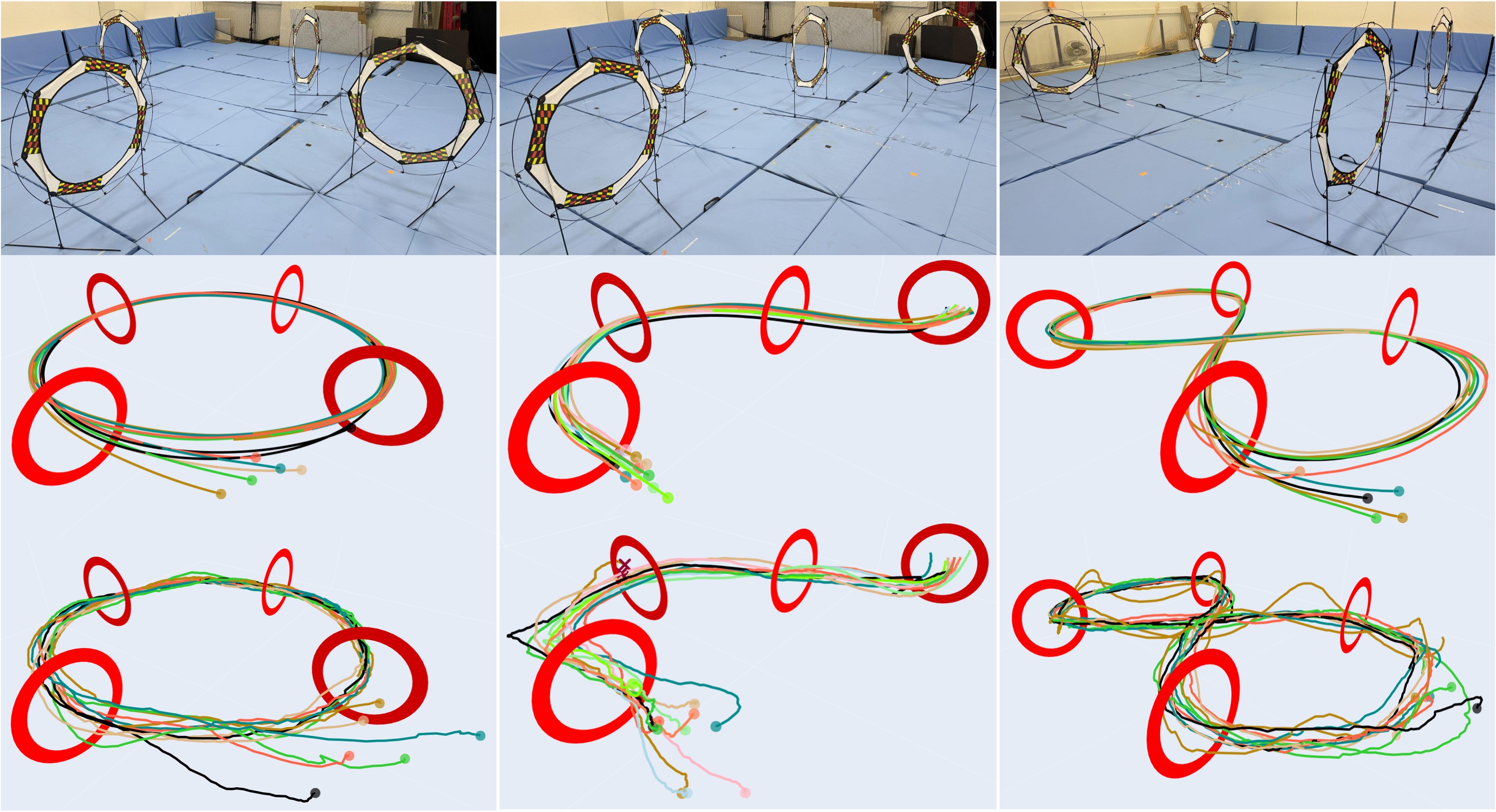}
    \caption{\small{Trajectory plots for three tracks: circle (left), U-turn (middle), and figure-8 (right). The top row shows the real-world gate setups in our indoor flight area, the middle row visualize 10-lap trajectories in FalconGym using our Multi-Modal(MM) controller, and the bottom row displays trajectories from real hardware experiments, where the same controller was deployed without additional fine-tuning. In FalconGym, our MM controller achieves a 100\% gate crossing success rate across all tracks. Although real-world trajectories exhibit greater variability, the MM controller still achieves success rates of 100\%, 87.5\%, and 100\% on the circle, U-turn, and figure-8 tracks, respectively. Red crosses indicate points where the quadrotor collided with a gate.}
}
    \label{fig:trajactory_plot}
\end{figure*}

We define a \emph{track} as a sequence of gates arranged in various poses within a designated flight area, as illustrated in Figure \ref{fig:trajactory_plot}. For each track, we want to build a FalconGym $F$ that can render a photorealistic image $v$ from any virtual camera pose $p$, i.e., $v = F(p)$. 

To train our FalconGym simulation environment, we first collect a dataset $\mathcal{D}_F = \{ (v_i, p_i) \}_{i=1}^{N}$ where $v_i \in \mathbb{R}^{w \times h \times 3}$ is an RGB image taken at camera pose $p_i\in \mathbb{R}^6$, which comprises 3D position and orientation. 
A human operator walks around each track for approximately five minutes, capturing $ N \sim 1500$ images from diverse viewpoints with an ArduCam RGB camera of known intrinsic parameters. 
Simultaneously, a localization marker is mounted on top of the camera and tracked by a Vicon\textsuperscript{\textregistered} motion capture system.
This system yields sub-millimeter precision ($\approx0.15mm$ mean error ~\cite{merriaux:hal-01710399}) and eliminates the need for Structure from Motion (SfM) pipelines such as COLMAP \cite{7780814}, which typically introduce larger alignment and scale uncertainties than a motion capture system.
All sensor data, including camera images and pose information, are synchronized via ROS.

After approximately 15 minutes of training $\mathcal{D}_F$ with NeRF \cite{10.1145/3503250} on an NVIDIA RTX~4090 using the open-source nerfstudio library~\cite{nerfstudio}, we obtain a FalconGym model capable of generating photorealistic images for any novel camera poses. Figure~\ref{fig:real_vs_sim} provides a qualitative comparison between real-world images and those generated by FalconGym.

% This high-fidelity FalconGym environment facilitates subsequent perception tasks and vision-based controller training entirely in simulation.

\subsection{Perception Module: NPE with Kalman Filter}
\label{sec:perception-module}

State estimation is essential for stable quadrotor flight, as it provides crucial pose feedback needed by the controller. 
In this work, we adopt a standard RGB camera (Arducam), popular in quadrotor setups, rather than more expensive or computationally demanding tracking cameras or LiDARs.

To estimate camera pose directly from an RGB image, we propose a \emph{Neural Pose Estimator (NPE)}, as shown in the blue box in Figure \ref{fig:perception-control-architecture}.
An iterative approach like iNeRF~\cite{yen2020inerf} can also estimate camera pose by optimizing the pose to match a target image through gradient-based refinement, but that process is too slow for real-time flight. 
Instead, we train a single-shot neural network NPE to estimate pose.

For the backbone of NPE, we use a pretrained Vision Transformer (ViT)~\cite{dosovitskiy2021an} on ImageNet-21k~\cite{ridnik2021imagenetk}, freezing its early layers to leverage robust, generic feature representations, then we attach a learnable regression head to predict camera pose. 
Instead of directly regressing orientation (pitch, roll, and yaw), we predict their sine and cosine values to avoid discontinuities associated with circular data at $2\pi$, and later converted to orientation using a trigonometric transformation.

A key advantage of neural scene representation is the ability to generate large synthetic datasets without the overhead of real-world data collection. For each of our \emph{FalconGym} simulation environment, we randomly sample $M \sim 10$k camera poses $p_j$ in the workspace and render corresponding images $v_j = F(p_j)$ using the previously trained FalconGym to form a dataset $\mathcal{D}_{N} = \{(v_j, p_j) \}_{j=1}^M$. We then train NPE to learn the mapping $v \mapsto p$.

Despite this rich synthetic dataset, single-frame predictions may still be noisy under challenging visual conditions (e.g., motion blur or ambiguous viewpoints), leading to abrupt frame-to-frame jumps.
To mitigate such transient errors, we fuse the NPE estimates with IMU data $z \in \mathbb{R}^6$ (representing three-dimensional linear and angular accelerations) in a Kalman filter, assuming a double integrator model for the quadrotor’s dynamics. 
% which is valid for low-speed flights ($<1.5\,\mathrm{m/s}$) based on our empirical testing. 
% 
This fusion strategy not only smooths out pose estimates, but also could accommodate asynchronous sensor rates in hardware experiments: if the camera runs at a lower frame rate, the Kalman filter can still update state estimates from the IMU until the next image arrives. 

We model the entire perception pipeline, which consists of NPE followed by the Kalman filter, as a function $P$; that is, given an input image $v$ and IMU data $z$, the model outputs an estimated pose $\hat{p} = P(v, z)$. 
In practice, we observe that it takes about 1 hour to train the NPE model, and the entire perception pipeline has an average of $7ms$ inference time on a NVIDIA RTX~4090, making the proposed perception pipeline feasible for real-world deployment.

% In the next section, we detail how these refined state estimates guide the design of our controller architecture.

\subsection{Controller Architecture: Self-Attention}
\label{sec:controller-architecture}

Although one could directly feed the estimated state to a state-based controller (as explored in Section~\ref{sec:cotrol_eval}), large perception errors could affect downstream controller and lead to constraints violations like collision or off-track.
Inspired by residual connections in deep networks \cite{he2016residual} that can improve model accuracies and attention mechanism \cite{10.5555/3295222.3295349} in fusing different inputs, we instead design a Multi-Modal (MM) controller that fuses visual feature and state information with self-attention, as illustrated in the purple box of Figure \ref{fig:perception-control-architecture}.

For the visual branch of the MM controller, we employ another pre-trained, partially frozen ViT \cite{dosovitskiy2021an} to process raw RGB images. 
The output of its final layer is passed through a trainable head that projects the visual features into a 256-dimensional embedding, capturing essential gate-related vision cues. 
In parallel, on the state branch of the MM controller, we feed the estimated state (from Section~\ref{sec:perception-module}), IMU data, and the next gate's ground truth pose (obtained when we arranged the track) into a Fully Connected (FC) network, obtaining a second 256-dimensional embedding.

To effectively leverage the visual and state modalities, we apply a multi-head self-attention layer to the concatenated visual and state embeddings. 
This mechanism enables the network to adaptively focus on whichever branch (visual or state-based) is most relevant in a given context. 
For instance, when the gate is clearly visible, the attention mechanism can emphasize image features, whereas in poor gate visibility or occlusion scenarios, it can learn to rely more on IMU data. 
Finally, another trainable FC network outputs the control $u \in \mathbb{R}^3$ (representing the acceleration in three directions), which are then fed into closed-loop control in both FalconGym simulation environments and real-world experiments.

\begin{figure*}[htbp]
    \centering
    \includegraphics[width=\linewidth]{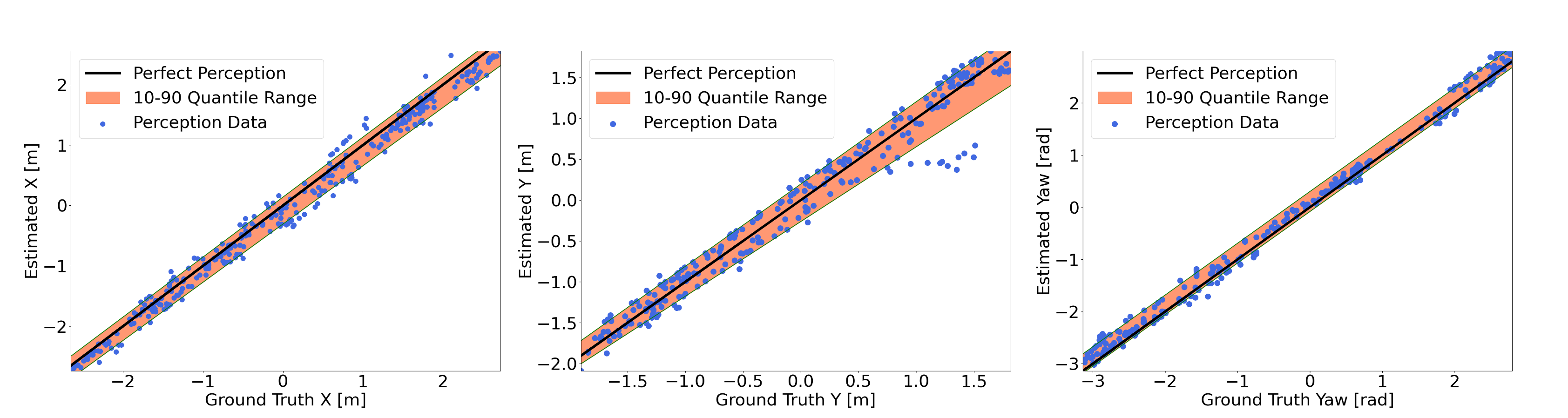}
    \caption{
    \small{Qualitative evaluation of the perception module \(P\) is shown. Blue dots represent a subset of $\mathcal{D}_Q$ (see Section~\ref{sec:imitation-learning}), which compare the pose estimation from perception module $P$ against the ground truth pose. The black line indicates perfect estimation, while the orange shaded area denotes the 10–90\% quantile range of perception errors, which is used to augment the controller training dataset \(\mathcal{D}_C\). The close clustering of tblue dots around the black line demonstrates that our perception module achieves accurate pose estimation.}
}
    \label{fig:perception-analysis}
\end{figure*}

Yaw is handled separately by a proportional controller, since the yaw estimates from our perception module is relatively accurate (see Table \ref{tab:perception-result} and Figure \ref{fig:perception-analysis}). 

% Overall, by sensor fusing visual embeddings with state embeddings through self-attention, our controller adaptively balances reliance on each modality, and enhances the gate-navigation success rate, as we will show in Section \ref{sec:experiment}.

\subsection{Training via Imitation Learning (DAGGER)}
\label{sec:imitation-learning}

Our objective is to train the Multi-Modal (MM) controller with self-attention mechanism to cross gates on various tracks, by learning from a teacher in a DAGGER-style imitation learning framework \cite{pmlr-v15-ross11a}.

\subsubsection{Expert Controller}
We begin by implementing an expert state-based controller \cite{Sun2020} that uses precise ground-truth states and gate poses to produce expert actions for gate navigation, i.e. $u^*_t = \pi^*(p_t, z_t, g_t)$, 
where $p_t$ is the ground-truth camera (quadrotor) pose, $z_t$ is the IMU data, $g_t \in \mathbb{R}^6$ is the next gate center’s pose, and $u^*_t$ is the expert acceleration control generated by the expert controller $\pi^*$ at timestep $t$.
% Although highly effective, this expert relies on perfect state information and thus does not constitute a vision-based solution. Nevertheless, it serves as the teacher for our DAGGER algorithm. 

During each iteration, we inject carefully-designed mild random noise into expert's control output, i.e. $u_t'^{*} = u_t^* + \mathcal{N}(0, \sigma^2)$, to encourage the quadrotor to visit states beyond its nominal trajectory, yet still remain capable of crossing gates.
% $u^{*\prime}_t = u^*_t + \delta_t $ where $\delta_t \sim \mathcal{N}(0,\sigma^2)$.
This practice, akin to domain randomization~\cite{doi:10.1126/scirobotics.abg5810} and DAGGER \cite{pmlr-v15-ross11a}, broadens the data distribution and exposes the policy to less-frequently visited states.

\subsubsection{Quantile-Based Perception Error Modeling}
Inspired by the ``perception contract'' concept~\cite{li2023refiningperceptioncontractscase,10.1109/TCAD.2022.3197508}, we explicitly model the error between ground-truth and estimated states. 
First, we employ the expert controller (with domain randomized control) in FalconGym, to collect a dataset $\mathcal{D}_Q = \{(\hat{p}_t, p_t)\}_{t=1}^{K}$ where $\hat{p}_t = P(F(p_t), z_t)$ is the pose estimation by NPE and Kalman filter (Section \ref{sec:perception-module}). We gathered $K \sim 10$k over 500 runs in FalconGym. 
Next, we apply quantile regression~\cite{f3e58db4-5840-30e3-a829-a85f9090df7a} to $\mathcal{D}_Q$ to estimate the conditional 10th and 90th percentiles of the estimation error distribution, denoted $Q_{0.1}(p)$ and $Q_{0.9}(p)$. 
These quantiles define an empirical 80\% prediction interval for the error term $\hat{p}-p$ such that 
\[
\mathbb{P}\Bigr(Q_{0.1}(p)\leq \, \hat{p}-p \, \leq Q_{0.9}(p) \; | \;p\Bigr)\geq 0.8
\]
as demonstrated experimentally in Figure~\ref{fig:perception-analysis}.

% We then fit a quantile regression model \cite{f3e58db4-5840-30e3-a829-a85f9090df7a} on $\mathcal{D}_Q$ to obtain the 10\% and 90\% quantiles, $Q_{0.1}$ and $Q_{0.9}$, which emphirically bound the estimation error $ \hat{p} - p \in [Q_{0.1}(p), Q_{0.9}(p)]$,as shown in Figure~\ref{fig:perception-analysis}.

Finally, rather than assigning a single estimated pose to each ground-truth pose, for each $p$, we randomly sample 50 \emph{perturbed} pose \(\tilde{p}\) within this interval:
$$
\tilde{p} = p +\text{Unif}\Bigr(Q_{0.1}(p),\,Q_{0.9}(p)\Bigr).
$$
The rationale is that, for a given true pose \(p\), multiple plausible estimates \(\hat{p}\) might exist, yet all of which should lead to the same expert action.
By data augmentation using this quantile range, we prepare our controller for perception uncertainties.

\subsubsection{Controller Dataset Collection.}
By above augmentation, we form the final controller dataset with size of $R \sim$500k samples:
$
\mathcal{D}_C = \bigl\{\,(v_t, \tilde{p}_t, z_t, g_t), (u^*_t))\bigr\}_{t=1}^{R}.
$
where $v_t = F(p_t)$ is FalconGym-rendered image at ground-truth pose $p_t$, $\tilde{p}_t$ is the sampled pose estimation from the above quantile range, $z_t$ is IMU data, $g_t$ is the next gate’s pose, and $u^*_t = \pi^*(p_t, z_t, g_t)$ is the expert action.

% To ensure broad coverage, we sample 50 distinct $\tilde{p}_t$ values for each $p_t$, increasing the dataset size to around $R \sim$500k samples. 
% Crucially, the expert command $u^*_t$ depends solely on the ground-truth state $p_t$; thus, minor perturbations in $\tilde{p}_t$ do not alter the correct action labels.

\subsubsection{Multi-Modal Controller Learning}
We train our MM controller $\pi(v_t,\tilde{p}_t,z_t,g_t)$ with supervised learning by minimizing the squared error between its predicted acceleration and the expert action:
\[
\min_{\pi} \sum_{t=1}^R \Bigl\|\hat{u}_t \;-\; u^*_t\Bigr\|^2,
\quad \text{where} \quad \hat{u}_t = \pi\!\bigl(v_t,\,\tilde{p}_t,\,z_t,\,g_t\bigr).
\]
The training takes about 10 hours on a 4090 GPU to finish due to large amount of training data, but the inference time of the controller is only about 11ms. 

% By combining DAGGER with quantile-based error sampling, our approach prepares the SF controller $p$ for a variaty of perception uncertainties it will encounter in real flight.

\begin{table*}[htbp]\centering
    \caption{Evaluation of the perception module. The table compares the Neural Pose Estimator (NPE) used alone versus with Kalman filtering, and evaluates perception module's performance of zero-shot sim-to-real transfer.}
    \label{tab:perception-result}

    \scriptsize
    \begin{tabular}{lcccc|cccc}\toprule
        &\multicolumn{4}{c|}{FalconGym} &\multicolumn{4}{c}{Real World} \\\cmidrule{2-9}
        &\multicolumn{2}{c}{NPE} &\multicolumn{2}{c|}{NPE + Kalman Filter} &\multicolumn{2}{c}{NPE} &\multicolumn{2}{c}{NPE + Kalman Filter} \\\cmidrule{2-9}
        
        % &\multicolumn{2}{c}{FalconGym: NPE} &\multicolumn{2}{c|}{FalconGym: NPE + Kalman Filter} &\multicolumn{2}{c}{Real World: NPE} &\multicolumn{2}{c}{Real World: NPE + Kalman Filter} \\\cmidrule{2-9}
        Track & Pose Err. [cm]$\downarrow$ & Yaw Err. [deg]$\downarrow$ & Pose Err. [cm]$\downarrow$ & Yaw Err. [deg]$\downarrow$ & Pose Err. [cm]$\downarrow$ & Yaw Err. [deg]$\downarrow$ & Pose Err. [cm]$\downarrow$ & Yaw Err. [deg]$\downarrow$
        % \\
        % & [cm] & [degrees] & [cm] & [degrees]& [cm] & [degrees]& [cm] & [degrees]
        \\\midrule
        Circle & 33.9 & 4.7 & \textbf{21.8} & \textbf{4.1} & 34.9  & 6.3 & \textbf{22.1} & \textbf{6.0} \\
        U-turn & 55.5 & 11.0 & \textbf{32.3} & \textbf{7.6} & 47.5 & 19.8 & \textbf{33.6} & \textbf{12.6} \\
        Figure-8 & 42.4 & 7.7 & \textbf{24.0} & \textbf{6.4} & 44.3 & 12.4 & \textbf{26.2} & \textbf{9.0}\\
    \bottomrule
    \end{tabular}
\end{table*}

\section{EXPERIMENTS}

\label{sec:experiment}

In this section, we first describe our hardware setup and track configuration. 
We then evaluate the proposed approach in the previous section using well-established performance metrics and compare it with state-of-the-art baseline.

\subsection{Closed-Loop System Setup in Hardware}

We conduct our hardware experiments on a quadrotor measuring 30cm $\times$ 34cm $\times$ 11cm, equipped with a Raspberry~Pi~3\textsuperscript{\textregistered} and a Navio2\textsuperscript{\textregistered} board for flight control, as shown in Figure \ref{fig:drone_gates}.
The Navio2\textsuperscript{\textregistered} incorporates an onboard IMU and streams acceleration data at up to 50hz. 
An ArduCam global-shutter camera is also mounted on the quadrotor, in order to reduce both computation and communication overhead, we operate the camera at 20Hz and downsample the image size to $320\times240$.

All sensor data, including camera images and IMU measurements, are published to ROS topics on the Pi. 
An offboard workstation, connected on the same subnet as the Pi and equipped with an NVIDIA RTX~4090 GPU, subscribes to these topics.
Then we perform the perception (Section~\ref{sec:perception-module}) and controller inference (Section~\ref{sec:controller-architecture}) offboard, and finally sends the computed control actions back to the Pi via another ROS topic. 
Empirically, we measure an end-to-end latency of about 39ms between when sensor data are sent and when control commands are received. 
This allows us to maintain a stable 20Hz (50ms delay) control loop in the hardware. To be consistent, we also use a $\Delta t = 0.05s$ step size (i.e. 20Hz) for training in FalconGym as well.

Our indoor flight area measures 6m $\times$ 6m $\times$ 3m, given this limited operational area, the expert controller is configured to fly at a constant velocity of 1m/s for safety reason. 
Empirically, we observe that at this speed, the quadrotor's motion closely follows a double integrator model, justifying its use as the assumed dynamics in FalconGym. 
We design three distinct tracks within the flight area, each comprising four gates with 38 cm inner radius, of diverse geometries and sharp turns (namely circle, figure-8, and an acyclic U-turn), as shown in Figure \ref{fig:trajactory_plot}. During track setup, we pre-record the precise poses of all gates to ensure accurate gate information for both simulation and hardware experiments.

Although a Vicon\textsuperscript{\textregistered} motion-capture system is available, as noted in Section~\ref{sec:nerf-construction}, it is used strictly for logging the quadrotor’s trajectories and identifying the next gate center based on the quadrotor’s live position. 
The next gate's pose information is relayed to the offboard workstation through ROS as another input to the MM controller, as shown in Figure \ref{fig:perception-control-architecture}, but our controller does not use any ground truth quadrotor pose information from Vicon\textsuperscript{\textregistered}. 

\subsection{Validation of FalconGym}
In Figure \ref{fig:real_vs_sim}, we qualitatively demonstrated the realistic visual fidelity of our FalconGym. 
We now provide a quantitative analysis using three standard metrics for neural scene representation: peak signal-to-noise ratio (PSNR) which measures the pixel-wise reconstruction quality, structural similarity index (SSIM) \cite{1284395} which gauges structural similarity, and learned perceptual image patch similarity (LPIPS) \cite{8578166} that captures perceptual differences by comparing feature embeddings.

For each track, we compare hundreds of the real ArduCam-captured images against the images generated by our FalconGym. 
As summarized in Table \ref{tab:nerf-metrics}, the reconstructed scenes achieve an average PSNR of 28.3\,dB, SSIM of 0.87, and LPIPS of 0.27 across the three tracks. 
These values are consistent with those reported by the original NeRF work on a real dataset \cite{10.1145/3503250}. 
Consequently, our FalconGym serves as a solid foundation for the subsequent perception and control modules.

\subsection{Perception Module Evaluation}
Next, we evaluate the perception module from Section~\ref{sec:perception-module} to answer two main questions: (i) Does Kalman filtering reduce NPE estimation error? (ii) Can a perception model trained entirely in our FalconGym maintain similar performance when transferred to real world?

We measure two key error metrics: the error between the predicted and ground-truth pose, and the error between the predicted and ground-truth yaw. 
% Since yaw alignment is critical for navigating gates, in this work, we focus on yaw and overall positional accuracy while ignoring minor pitch/roll variations.
To isolate perception performance from the controller, we use the expert state-based controller \cite{Sun2020} to fly 10 laps on each track from slightly different initial poses, both in FalconGym and in hardware experiments. 
Over these trials, we record ground-truth poses, raw NPE estimates and Kalman-filtered estimates. 

Table \ref{tab:perception-result} summarizes these perception results. We see that, in FalconGym, the NPE alone achieves relatively accurate yaw prediction. 
However, NPE's position error in FalconGym averages 43.9cm, which poses a risk for gates of only 38cm radius.
Once fused with the Kalman filter, we see a significant drop in pose estimation error (26\,cm), given that IMU data compensate for some perception error.
Comparing FalconGym and real world data, we observe slightly higher perception errors in the real-world setting (an average of 27.3cm) than FalconGym (an average of 26cm), likely reflecting minor domain gaps between rendered FalconGym images and actual camera feeds.

Overall, these findings confirm that Kalman filter enhances state estimation and the overall perception module has a similar sim-to-real performance.

\begin{table}[htbp]\centering
\caption{Evaluation of FalconGym Photorealism quality.
}
\label{tab:nerf-metrics}
\scriptsize
\begin{tabular}{lcccccc}\toprule
    \multicolumn{1}{l}{Track}
    &\multicolumn{1}{c}{\# Images}
    &\multicolumn{1}{c}{PSNR(dB)$\uparrow$} 
    &\multicolumn{1}{c}{SSIM$\uparrow$} 
    &\multicolumn{1}{c}{LPIPS$\downarrow$} 
    &\multicolumn{1}{c}{Time} 
    \\\midrule
    Circle & 1583 & 28.3694 & 0.8720 & 0.2689 & 12m 49s \\ 
    U-turn & 1094 & 28.3205 & 0.8677 & 0.2929 & 12m 33s\\
    Figure-8 & 839 & 28.3691 & 0.8852 & 0.2665 & 12m 48s\\
\bottomrule
\end{tabular}
\end{table}

% \begin{enumerate}
%     \item high success rate + low error
%     \item slightly beat baseline when gates are visible
%     \item dramatically beat baseline when gates are not visible for several frames
% \end{enumerate}

\subsection{Controller Evaluation}
\label{sec:cotrol_eval}
Following the visualized gate-crossing trajectories shown in Figure~\ref{fig:trajactory_plot}, we next present a quantitative analysis that addresses two main questions:(i)~Does our Multi-Modal (MM) controller outperform a vision-only state-of-the-art (SoTA) baseline? (ii)~Can our MM approach transfer effectively from FalconGym to real hardware?

We quantify how well the quadrotor controller upholds the hard constraints (avoiding collisions or going off-track) using two metrics: (i)~\emph{Success Rate (SR)} measures the percentage of gates that the quadrotor safely flies through; (ii) \emph{Mean Gate Error (MGE)} evaluates the average distance between the quadrotor and the gate center at the time of crossing.

We implemented four different controllers for our experiments: (i) Expert (State-based): A controller from Sun et~al.~\cite{Sun2020} with full ground-truth state access;  (ii) SoTA vision-only Baseline: A method adapted from Geles et~al.~\cite{DBLP:conf/rss/GelesBRX024}, where a privileged critic observes the full state during training, while the actor relies on images and past actions at inference and do not need IMU; (iii) Direct Perception (DP): our naive method that directly feeds our perception output from NPE and Kalman filter into the state-based controller; (iv) Multi-Modal(MM): our approach which fuses image embeddings and state embeddings via self-attention
% (Section~\ref{sec:controller-architecture}). 

To remain faithful to the baseline’s setup, we apply the same gate-detection method of ~\cite{DBLP:conf/rss/GelesBRX024} (HIL experiments) in simulation that projects known 3D gate positions into 2D image coordinates using the camera’s intrinsics and extrinsics. 
In addition, because the baseline struggles when gates disappear for consecutive frames, we further add a yaw-control heuristic to keep the quadrotor facing the next gate whenever possible.

Each of the fours controllers is run for 10 laps on the circle, U-turn, and figure-8 tracks (4 gates per track) with slightly varied initial poses in FalconGym.
DP and the baseline are excluded from the real world hardware experiments, as their low simulation success rates raise safety concerns.

\begin{table}[htbp]\centering
\caption{Evaluation of controllers by Success Rate (SR) and Mean Gate Error (MGE) in FalconGym and real world.}
\label{tab:controller-analysis}
\scriptsize
\begin{tabular}{lccc|cc}\toprule
    &  & \multicolumn{2}{c|}{FalconGym} & \multicolumn{2}{c}{Real World} \\\midrule
    & Method & SR $\uparrow$ & MGE [cm] $\downarrow$ & SR $\uparrow$ & MGE [cm] $\downarrow$ \\\midrule
    Circle & \underline{State-based} & \underline{100\%}  & \underline{2.49}  & \underline{100\%} & \underline{6.22} \\
    & Baseline \cite{DBLP:conf/rss/GelesBRX024} & 100\% & 6.62 & n/a & n/a \\
    & Ours (DP) & 90\%  & 16.0  & n/a & n/a \\
    & \textbf{Ours (MM)}& \textbf{100\%} & \textbf{6.25} & \textbf{100\%} & \textbf{10.7} \\\midrule
    
    U-turn  & \underline{State-based} & \underline{100\%} & \underline{3.42}  & \underline{100\%} & \underline{9.80}\\
    & Baseline \cite{DBLP:conf/rss/GelesBRX024} & 100\% & 11.9 & n/a & n/a \\
    & Ours (DP) & 52.5\% & 11.3 & n/a & n/a \\
    & \textbf{Ours (MM)} & \textbf{100\%} & \textbf{10.1} & \textbf{87.5\%} & \textbf{11.1}  \\\midrule
    
    Figure-8 & \underline{State-based} & \underline{100\%} & \underline{5.11} & \underline{100\%} & \underline{9.30}  \\
    & Baseline \cite{DBLP:conf/rss/GelesBRX024} & 25\% & 8.44 & n/a & n/a \\
    & Ours (DP) & 20\% & 16.2 & n/a & n/a \\
    & \textbf{Ours (MM)} & \textbf{100\%} & \textbf{5.13} & \textbf{100\%} & \textbf{9.40}\\\bottomrule
\end{tabular}
\end{table}

Table~\ref{tab:controller-analysis} summarizes the results. 
In FalconGym, we see that MM controller consistently achieves higher SR (100\%) and lower MGE than DP, primarily due to sensor fusion compensating for inaccurate perception. 
The baseline also attains 100\% SR on circle and U-turn tracks but has slightly worse MGE than our MM controller and experiences a drop in SR on the figure-8 track, where gates remain out of view for extended periods. 
By contrast, our MM controller maintains 100\% SR across all tracks. 
The expert controller achieves near-perfect MGE thanks to full ground-truth access, which naturally surpasses our MM controller learned by imitation learning.

In real world hardware experiments (right portion of Table~\ref{tab:controller-analysis}), the expert again demonstrates flawless gate crossings but shows a slight increase in MGE, likely due to discrepancies between the real quadrotor and the approximate double-integrator dynamics used in FalconGym. 
Our MM controller achieves 100\% SR on the circle and figure-8 tracks and 87.5\% SR on the U-turn track (5 failures out of 40 gates over 10 runs). 
Further investigation shows that these failures consistently occur at the second gate of the U-turn track. 
We identified a few minor artifacts in the corresponding FalconGym simulation environment that appear to introduce a non-trivial sim-to-real visual gap: when we provide the MM controller with an artifact-affected image instead of a real-world image—while keeping the camera pose and other inputs identical—we observe a marked difference in its control outputs.
% These results suggest that the visual discrepancies between FalconGym and the real world on the uturn track are the primary cause of the observed failures.
% 
Nonetheless, the MGE for all tracks remains sufficiently low (average of 10.4cm) relative to the 38\,cm gate radius, and only slightly higher than its FalconGym counterpart (average of 7.2cm).
 These results indicate that our controller effectively conforms to the hard constraints of avoiding collisions and staying within gate boundaries in real-world scenarios.

Overall, these experiments show that (i)~our MM controller surpasses both naive DP approach and vision-only SoTA baseline in FalconGym by effectively fusing sensor inputs through self-attention, and (ii)~our MM controller generalizes well from FalconGym to hardware, validating our approach's ability for zero-shot sim-to-real quadrotor crossing gates.

\section{CONCLUSION}

In this work, we present a novel visual control framework for quadrotor gate navigation that achieves zero-shot sim-to-real transfer. Leveraging our photorealistic FalconGym simulation environment, we train a visual controller that integrates visual features and pose estimation through a self-attention-based sensor fusion mechanism, making it robust to perception errors. 
Experiments in both FalconGym and hardware demonstrate high performance of our approach.
% Overall, this paper establishes a practical pathway for resilient, vision-based quadrotor flight through diverse gate configurations.

Nonetheless, performance can degrade due to visual domain gaps, mismatches between FalconGym’s simulated dynamics and the true quadrotor, and offboard computation or communication delays. Future directions include replacing our NeRF backend with Gaussian Splatting~\cite{Wu_2024_CVPR} for faster, higher-fidelity rendering; training policies that generalize to unseen tracks; incorporating physics-based dynamics~\cite{Bauersfeld__2021} beyond a double-integrator model for aggressive maneuvers; distilling perception and control modules for fully onboard execution; and exploring high-speed flights to handle motion blur and tight perception–action coupling.

\section*{ACKNOWLEDGMENT}
The authors conducting this research were supported in part by the Air Force Research Laboratories (Award number FA8651-24-1-0007) and the Boeing Company. The authors would like to thank Yangge Li and Ege Yuceel for their valuable feedback.

% \begin{thebibliography}{99}

\bibliographystyle{IEEEtran}
\bibliography{IEEEabrv,references}
% \end{thebibliography}

\end{document}